\documentclass{article}
\usepackage{cite}
 \PassOptionsToPackage{numbers, compress}{natbib}

\usepackage[preprint]{neurips_2021}

\usepackage[utf8]{inputenc} 
\usepackage[T1]{fontenc}    
\usepackage[colorlinks,linkcolor=red,citecolor=green]{hyperref}
\usepackage[doipre={doi:~}]{uri}
\usepackage{url}            
\usepackage{booktabs}       
\usepackage{amsfonts}       
\usepackage{nicefrac}       
\usepackage{microtype}      
\usepackage{xcolor}         
\usepackage{subfigure} 
\usepackage{times}
\usepackage{epsfig}
\usepackage{graphicx}
\usepackage{amsmath}
\usepackage{amssymb}
\usepackage{wrapfig}
\usepackage{algorithmic}
\usepackage{algorithm}
\usepackage{caption}
 \usepackage{subfigure}
\usepackage{multirow}
\usepackage{comment}

\newcommand{\Obj}{\mathcal{G}}
\newcommand{\Cost}{\mathcal{L}}

\title{You Only Compress Once: Towards Effective and Elastic BERT Compression via Exploit-Explore Stochastic Nature Gradient }

\author{%
  Shaokun Zhang\,$^{1*}$ \quad Xiawu Zheng\,$^{1,4}$\thanks{Equal Contribution.} \quad Chenyi Yang\,$^1$ \quad Yuchao Li\,$^2$ \\ 
  \textbf{Yan Wang\,$^5$ \quad Fei Chao\,$^1$ \quad Mengdi Wang\,$^2$ \quad Shen Li\,$^2$  \quad Yang Jun\,$^{2}$ \quad Rongrong Ji\,$^{1,3,4}$\thanks{Corresponding Author: rrji@xmu.edu.cn}}\\
  $^1$Media Analytics and Computing Lab, Department of Artificial Intelligence, \\School of Informatics, Xiamen University \;\;\;\;\;\; 
  $^2$Alibaba Group \\ $^3$Institute of Artificial Intelligence, Xiamen University\\
  $^4$Peng Cheng Laboratory, Shenzhen, China.\\
  $^5$Pinterest, USA \\
}

\begin{document}

\maketitle

\begin{abstract}
Despite superior performance on various natural language processing tasks, pre-trained models such as BERT are challenged by deploying on resource-constraint devices.
Most existing model compression approaches require re-compression or fine-tuning across diverse constraints to accommodate various hardware deployments. This practically limits the further application of model compression. Moreover, the ineffective training and searching process of existing elastic compression paradigms~\cite{hanruiwang2020hat,Cai2020Once-for-All:} prevents the direct migration to BERT compression. Motivated by the necessity of efficient inference across various constraints on BERT, we propose a novel approach, YOCO-BERT, to achieve \emph{compress once and deploy everywhere}.
%
Specifically, we first construct a huge search space with $10^{13}$ architectures, which covers nearly all configurations in BERT model. Then, we propose a novel stochastic nature gradient optimization method to guide the generation of optimal candidate architecture which could keep a balanced trade-off between explorations and exploitation.
When a certain resource constraint is given, a lightweight distribution optimization approach is utilized to obtain the optimal network for target deployment without fine-tuning.
Compared with state-of-the-art algorithms, YOCO-BERT provides more compact models, yet achieving 2.1$\%$-4.5$\%$ average accuracy improvement on the GLUE benchmark. Besides, YOCO-BERT is also more effective, \emph{e.g.,} the training complexity is $\mathcal{O}(1)$ for $N$ different devices. Code is available \url{https://github.com/MAC-AutoML/YOCO-BERT}.
\end{abstract}

\section{Introduction}
\label{sec:1}
While BERT~\cite{devlin2019bert} has driven advances in various natural language processing tasks~\cite{yang-etal-2019-end-end,nogueira2019passage}, it is still notoriously challenging to deploy on resource-limited devices, due to its large demands on computational power and memory footprint. To alleviate this issue, there has been an exponential increase in research concerning compressing the model, such as knowledge distillation \cite{sun2019patient,sanh2019distilbert,jiao-etal-2020-tinybert}, quantization \cite{shen2020q}, weight factorization \cite{lan2019albert} and pruning \cite{gordon2020compressing,McCarley2019PruningAB}. Given an expected budget on computational power or memory, these methods compress the full-fledged BERT into a smaller and/or faster model, with acceptable compromise on model performance.
While effective, these methods are not sufficiently scalable to accommodate a wide range of devices, as re-compression or an expensive fine-tuning process is usually required for even a small change on the budget. 
It is thus impractical to adapt BERT or similar Transformer-based models to a reasonably large number of types of devices.

Recently, there have been pioneer works aiming to improve the flexibility of BERT compression. DynaBERT~\cite{hou2020dynabert} compresses BERT into smaller BERT-like sub-networks with adaptive widths and heights by training them iteratively using distillation.
AdaBERT and NAS-BERT ~\cite{ijcai2020-341,xu2021taskagnostic} exploits a NAS method to compress the BERT model in a huge convolution-based space, to achieve more flexibility in architecture.
While these two directions achieve deployment flexibility by serving different sub-networks at runtime, if viewed from an explore-exploit perspective, they lie in the two extremes of the spectrum.
By constraining the architecture of sub-networks to be the same as the model to be compressed, DynaBERT focuses on exploitation with tweaks on the width and height of Transformers.
On the contrary, AdaBERT and NAS-BERT nearly completely discard the Transformer architecture but perform an architectural search in a general search space based on convolution.

However, neither of these extreme approaches is optimal.
While the pure exploration approaches are able to achieve more flexibility from the huge search space, conventional NAS methods tend to converge to local minimas, and thus result in sub-optimal model performance.
The pure exploiting approaches strictly stick to the prior design of Transformers, and the limited search space soon becomes the bottleneck of an effective trade-off between model size and accuracy.
We argue that a proper balance between exploitation and exploration is the key to effective BERT model compression.
And this can be achieved with a huge but dedicatedly designed architectural search space, which is then explored and exploited by an effective search algorithm.
Different from the traditional sub-network-based compression approaches \cite{hou2020dynabert,Cai2020Once-for-All:,hanruiwang2020hat}, which randomly sample (i.e. explore) and train sub-networks to update the weights in super models, we formulated the search algorithm in a probabilistic manner.
This is inspired by stochastic natural gradient methods~\cite{shirakawa2018dynamic, AkimotoICML2019}.
These methods formulate the NAS process, which is traditionally thought non-differentiable, as a continuous likelihood maximization process.
On one hand, this allows us to effectively explore the search space to achieve faster convergence.
On the other hand, we can also easily plug in different distributions to model exploitation.

Therefore, in this paper, we propose an efficient and elastic BERT compression algorithm, YOCO-BERT, to enable maximal flexibility across a wide range of devices and minimal performance compromise, thus to achieve \emph{compress once and deploy everywhere}. 
We first construct a huge search space with $10^{13}$ architectures. Comparing with Dynabert\cite{hou2020dynabert} only has a dozen options, our search space covers nearly all configurations in BERT model.
Then, a novel exploit-explore balanced stochastic natural gradient optimization algorithm is proposed to efficiently explore the search space.
Specifically, there are two sequential stages in YOCO-BERT. We decouple the compression process into a ``super mode'' training process, which does the core heavy lifting, and a lightweight model adaptation process, which directly inherits the weight from the ``super model'' \emph{without any finetuning process}.

In the Exploit-Explore Stochastic Natural Gradient (EE-SNG) optimization algorithm, a probability distribution group containing an exploration distribution and an exploitation distribution is introduced in the super-BERT training process to sample potential optimal sub-BERTs architectures at each training step. Joint optimizations are then adopted on the weight across architectures and the parameters of the exploitation distribution. Thus, the ``promising'' architectures will be more likely to be sampled. We further introduce an online learnable controller to dynamically determine whether exploitation or exploration is more important.
More specifically, the controller is determined by the information entropy of the exploitation distribution, \emph{i.e.,} the probability the exploitation is increasing with the information entropy. In the deployment stage, we propose a novel objective function to obtain the optimal architecture given any constraints based on the optimized distribution. 
Thanks to the EE-SNG in the training stage, weights of the optimal architectures under different constraints are well optimized. 
In other words, we do not need any fine-tuning process to construct the target model, which leads to an effective and elastic BERT compression.
\begin{figure*}[t]
\centering
\includegraphics[width=1.0\linewidth]{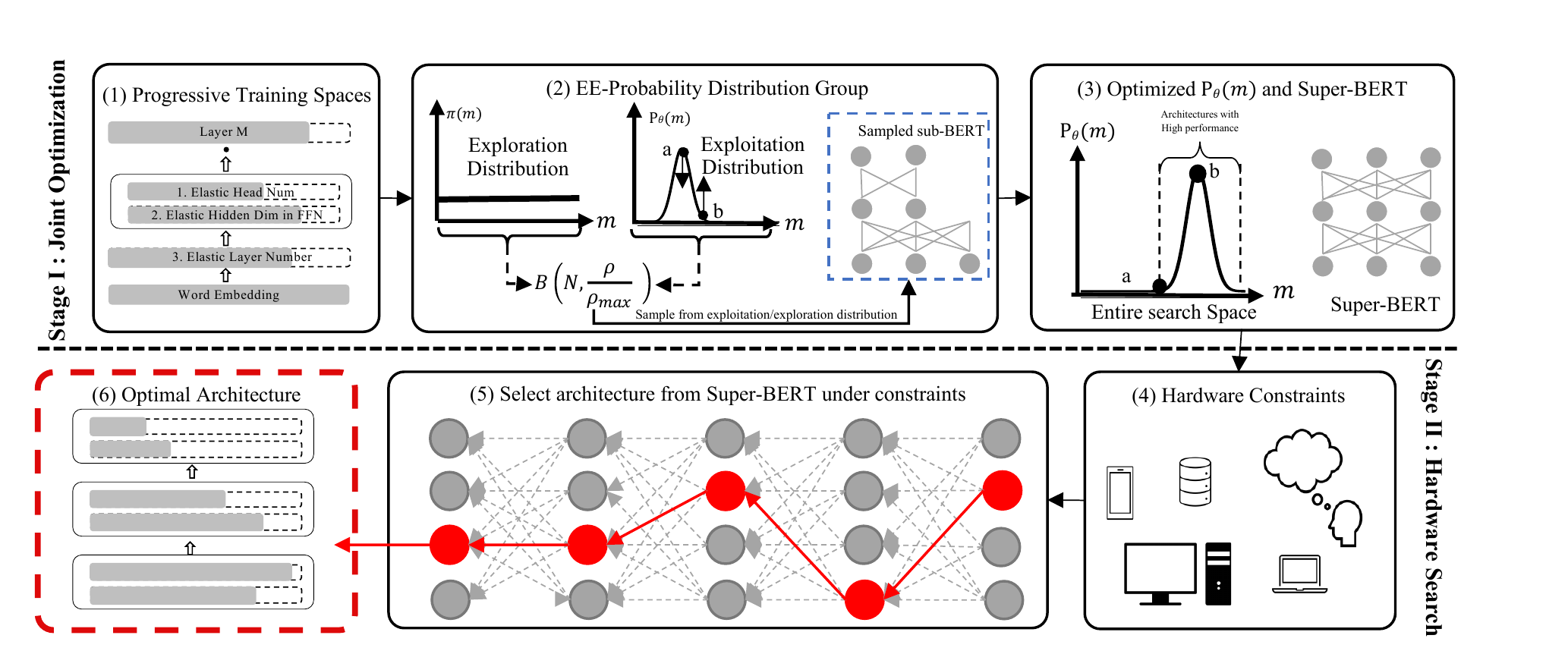}
\caption{The proposed YOCO-BERT. We separate YOCO-BERT into two stages. Given expanded training space, we first propose a EE-SNG to optimize the probability distribution $P_{\theta}$ and super-BERT. In the second stage, searching algorithm is supervised by the optimized $P_{\theta}$ to obtain the optimal architecture across diverse constraints. }
\label{fig:1}
\vspace{-0.2in}
\end{figure*}
The main contributions are listed as:

1) We design a novel and efficient BERT compression algorithm, YOCO-BERT, which enables scalability to a wide range of hardware.

2) We propose a novel stochastic nature gradient optimization method called EE-SNG to explicitly model exploitation and exploration in the model compression.

3) Comparing to previous methods \cite{sun2019patient,sanh2019distilbert,xu-etal-2020-bert,hou2020dynabert,fan2019reducing}, experiments show that our compressed models achieve the best compression rate and performance trade-off on GLUE\cite{wang2018glue} benchmark.

\section{Related Work}
Since most of the BERT compression methods have been discussed in Sec.~\ref{sec:1},  we introduce the two most related fields in this section.

\textbf{Neural Architecture Search.}
NAS technologies are proposed to automatically discover suitable network architectures over a vast architecture search space. 
Early works exploit the paradigms of reinforcement learning (RL)~\cite{45826} to reach that goal. 
However, the RL-based NAS methods require intensive computational and memory costs. \emph{e.g.,} \cite{45826} needs to train and evaluate more than $20,000$ neural architectures across $500$ GPUs over four days.
In order to alleviate this burden, one-shot NAS methods ~\cite{liu2018progressive,xie2020snas,liu2019darts,Cai2020Once-for-All:,zheng2019multinomial} proposed to construct a super-BERT that share the weights with all the sub-network in the search space. In this way, the search time will be reduced to few GPU hours and retaining a comparable accuracy since the sub-network could be evaluated efficiently. 

\textbf{Distribution Optimization.}
In terms of distribution optimization, various approaches have been proposed. For discrete optimization, there is population-based incremental learning (PBIL) \cite{baluja1994population}, and the compact genetic algorithm (cGA) \cite{harik1999compact}. Optimizing the parameters of the probability distribution is based on the IGO \cite{JMLR:v18:14-467}. 
There is a research direction of transforming NAS into a distribution optimization problem. Specifically, following \cite{shirakawa2018dynamic,AkimotoICML2019,9377468}, considering a probability distribution that generates network architecture and optimizing the parameters of the distribution instead of directly optimizing the network architecture. In this way, both the weight and the architecture will be differentiable which could be optimized simultaneously. However, such a method easily falls into local optimization due to reducing the probability of exploring the entire search space.

\section{YOCO-BERT}
In this section, we describe the proposed framework. Specifically, we first define the optimization problem in Sec.~\ref{sec:3.1}. Then, we discuss the search space in Sec.~\ref{sec:3.2}, the exploit-explore stochastic natural gradient in Sec.~\ref{sec:3.3} and the adaption of the optimized super-BERT for diverse constraints in Sec.~\ref{sec:3.4}. 

\subsection{Problem Definition}
\label{sec:3.1}
Our primary proposal is to optimize one super-BERT containing all sub-BERTs, enables the direct searching of candidate across diverse constraints. To this end, given a specific dataset, we address the following optimization problems:
\begin{equation}
\label{equ:1}
\begin{aligned}
     W^{*}= \arg\min_{W}\Cost_{train}\left(N\left(M,W\right)\right).& \\
\end{aligned}
\end{equation}
\begin{equation}
\label{equ:2}
\begin{aligned}
    &m^* = \mathop{\arg\max}_{m\in M}(\text{Eval}_\text{ACC}\left(m,W^{*}\left(m\right)\right), \, s.t. \; \mathcal{T}\left(m\right) < {\Omega}.&
\end{aligned}
\end{equation}
As demonstrated in the above equations, the problem is separated into two aspects.  Specifically, Eq.~\ref{equ:1} optimized the weight $W$ of the super-BERT $N\left(M,W\right)$ to minimize the training loss $\Cost_{train}\left(.\right)$ of all possible candidate architectures in search space $M$. The candidate architecture is denoted as $m\in M$. After the super-BERT is well optimized, the optimal architecture $m^*$ with satisfied constraint could be directly searched with weight $W^{*}\left(m\right)$ inherited from super-BERT. In Eq.~\ref{equ:2}, $\mathcal{T}\left(m\right)$ denotes the constraints for sub-BERT $m$ and $\Omega$ denotes the target constraints given for any devices. 
Note that, the optimal sub-BERT architecture $m^*$ directly inherited the weight from super-BERT to acquire the performance without any fine-tuning process, demonstrated in Tab.~\ref{tab:5}. Eq.~\ref{equ:1} and Eq.~\ref{equ:2} is solved by two corresponding stages, which are summarized as follows:

\textbf{Stage \uppercase\expandafter{\romannumeral1}: Joint Optimization} (Sec.~\ref{sec:3.3}).
We first solve the optimization problem of super-BERT (Eq.~\ref{equ:1}) in an optimization of a differentiable objective $\Obj(\cdot)$, the expectation of the loss on the training set. Formally, 
\begin{equation}
\label{equ:3}
\Obj \left(W, \theta\right) = E_{m \sim \mathcal{P}_{\theta}}\left(\Cost \left(N\left(w, m\right)\right)\right),
\end{equation}
where the function $\Cost \left(N\left(w, m\right)\right))$ is defined as the loss of the sub-BERT $N\left(w,m\right)$ on training set.
To this end, the search space is considered as a probability distribution $\mathcal{P}_{\theta}$ , which is parametrized by the hyper-parameter $\theta$. The expected loss $\Obj\left(\cdot\right)$ is minimized through optimizing the hyper-parameter $\theta$ and the weight $W$ of the super-BERT model.
In this paper, we propose a new Exploit-Explore Stochastic Natural Gradient (EE-SNG) method to optimize the aforementioned distribution. 

\textbf{Stage \uppercase\expandafter{\romannumeral2}: Hardware Search} (Sec.~\ref{sec:3.4}).
After the weight of super-BERT and the distribution are well optimized, the optimal sub-BERT could be directly obtained from super-BERT under specific constraints ${\Omega}$. This process is formally expressed in Eq.~\ref{equ:2}. In our algorithm, an efficient searching process could be achieved since the searching process is supervised by optimized distribution $\mathcal{P}_{\theta}$. 

\subsection{Search Space}
\label{sec:3.2}
\textbf{Search Space Design}.
YOCO-BERT automatically searches optimal transformer based architectures for BERT compression. In contrast to the convolutional neural networks, the transformer-based architecture consists of a word embedding layer and several transformer encoder layers, while each encoder layer is fully characterized by a multi-head self-attention mechanism and a position-wise feed-forward network. Instead of providing limited options of architecture settings~\cite{hou2020dynabert}, we specially design a more general search space for BERT, covering three main dimensions, \emph{i.e.}, encoder layers number $d$, intermediate size for two feed-forward layers $k$ and heads for each multi-head attention module $h$. In this way, we built a huge search space with $10^{13}$ architectures, allowing elastic head number, intermediate size, layer number to accommodate almost all hardware constraints.

\textbf{Super-BERT Model.}
To bypass the repeated training of candidate architectures, we train a parameters-shared network, super-BERT model, subsuming all sub-BERT models. Specifically, the weight of all candidate sub-BERT models are directly inherited from the front portion of its corresponding part in super-BERT after training super-BERT once. In this way, the optimized super-BERT is determined as an evaluation proxy for all sub-BERTs. Intuitively, the introduce of super-BERT model successfully decouple the parameters-shared super-BERT training and the candidate architecture search which greatly reduces the cost of sub-BERTs evaluation. Moreover, this decoupling enables to the acquisition of the sub-BERTs by extracting corresponding weights without any fine-tuning.

\textbf{Progressive Expanding Training Space}.
Inspired by the success of OFA~\cite{Cai2020Once-for-All:} in CNN-based space, we propose a progressive expanding training space in transformer-based space to reduce the interference between sub-BERTs. Specifically, the above dimensions are progressively expanded to train the super-BERT. We first train the largest candidate, $\text{BERT}_{\text{base}}$ model ($d = 12$, $k = 3,072$ and $h = 12$). Then, progressively shrink this candidate to support the small architectures to generate an elastic training space in each dimension (\emph{i.e.,} attention head number, intermediate size for two feed-forward layers and layer number). Specifically, the head numbers in each transformer encoder layer are relaxed to optional dimensions from a fixed size in step-by-step processing, then the same relaxation is applied for the intermediate size and the layer number. (\emph{i.e.,} the optional dimension of the head number is expanded from a fixed $\{12\}$ to $\{12,8\}$, and then expanded to $\{12,8,4\}$ at each layer. In each training step, all sub-BERTs are sampled from the exploit-explore probability distribution group of the expanding training space to optimize the corresponding part in the super-BERT.

\subsection{Exploit-Explore Stochastic Natural Gradient Optimization}
\label{sec:3.3}
In this section, we demonstrate the optimization of probability distribution $\mathcal{P}_{\theta}$ in Eq.~\ref{equ:3} and how to alleviate the dilemma of exploitation and exploration. Our goal is to minimize the loss expect of all candidates in training space, and thus the optimal architectures are more likely to be directly sampled from the optimized distribution. To this end, we formulate the training space as a probability distribution $\mathcal{P}_{\theta}$. Intuitively, we introduce a family of multi-dimensional probability distribution $P_{\theta}\left(m\right)$ defined on the training search space where $m$ denotes an sub-BERT architecture. Each sub-BERTs $m =(o,d)$ could be defined by dimensions selections of each operations. Intuitively, every architecture has a corresponding probability $P_{\theta}(m)$. The candidate architectures are obtained through a sequential process:  $\mathcal{P}_{\theta}\left(m\right) = p_{\theta_1}\left(o_{1}\right) \times p_{\theta_2}(o_{2}) ... p_{\theta_{i}}(o_{i})$. Notice that, different architecture variables are independent from each other. $o_{i}$ denotes an one-hot vector that represents the dimension selection of operation $i$.
Based on previous works~\cite{shirakawa2018dynamic,AkimotoICML2019}, the expectation of loss is differentiable for $w$ and ${\theta}$,
thus the distribution $\mathcal{P}_{\theta}$ could be optimized using \emph{stochastic relaxation}:
\begin{equation}\label{equ:4}
\begin{aligned}
&\nabla_w \Obj (W, \theta) = \mathbb{E}_{m \sim \mathcal{P}_{\theta}}[\nabla_w\Cost (w, m)], 
\\
&\tilde{\nabla_{\theta}} \Obj (W, \theta) = \mathbb{E}_{m \sim \mathcal{P}_{\theta}}[\Cost (w, m) \nabla_{\theta} \ln(P_{\theta}(m))].
\end{aligned}
\end{equation} 
Where $\Obj (W, \theta)$ denotes the expected loss for super-BERT with weight $W$ and distribution with hyper-parameter $\theta$. The $\tilde{\nabla}_{\theta} \ln(P_{\theta_i}(m)) = F_{-1}\ln(P_{\theta_i}(m))$ is the so-called natural gradient of the log-likelihood introduced in \cite{amari1998natural}. In practice, the gradients are estimated by the Monte-Carlo with samples $m^{i} \sim \mathcal{P}_{\theta}(i = 1,...,\lambda)$.

\begin{equation}
\label{equ:5}
\begin{aligned}
&\nabla_w \Obj (W, \theta) = \frac{1}{\lambda} \sum_{i = 1}^{\lambda}\nabla_w \Cost(w^i, m^i), 
\\
&\tilde{\nabla_{\theta}} \Obj (W, \theta) = \frac{1}{\lambda} \sum_{i = 1}^{\lambda} \Cost(w^{i}, m^i)\tilde{\nabla}_{\theta}\ln\mathcal{P}_{\theta}(m^{i}).
\end{aligned}
\end{equation} 
With sufficient statistics $U(m)$, the expectation parameterization $\theta=E(U(m))$ admits the natural gradient of the log-likelihood $\tilde{\nabla}_{\theta} \ln(P_{\theta}(m))=U(x) - \theta$. The weight $W$ and the parameters of the distribution $\theta$ will be optimized with appropriate learning-rates.

However, the greedy nature of the SNG~\cite{shirakawa2018dynamic} based methods inevitably introduces bias to ${\theta}$ and w during optimization, result a dilemma of exploitation and exploration. Intuitively, this bias leave a portion of architectures well trained and thus sampling more times, gradually leading a more serious biased sampling. Reduced probability of exploring the entire search space limited the upper bound of our optimization.
Therefore, we propose exploit-explore probability distribution group (EE-probability distribution group) to alleviate this issue, which is presented in Alg.~\ref{alg:1}. In the following, we discuss the collaboration of exploration and exploitation distribution with an online-learnable controller. 

\textbf{Exploit-Explore Probability Distribution Group.} We propose a probability distribution group to merge the benefits of both deterministic and stochastic predictions, significantly improving efficiency and effectiveness. The main distribution, the exploitation distribution $\mathcal{P}_{\theta}$, is optimized during the whole process. The parameters $\theta$ of the exploitation distribution and the weight $w$ of the sampled architectures are both optimized with sampled architectures using the gradient as shown in Eq.~\ref{equ:5}. However, such an optimization algorithm easily falls into a local optimum as state earlier. 
Specifically, we introduce a novel distribution $\pi$ defined on the whole search space. $\pi$ is always a uniform distribution which is considered as an exploration distribution. When the further optimization of $P_{\theta}$ is constraint by the aforementioned bias, the samplings from the distribution $\pi$ will supervise the optimization to explore the architectures which have a small probability sampled by $P_{\theta}$. In this way, the exploitation distribution avoids falling into local optima since introducing the uncertainty and randomness from the entire search space. A critical problem of the explore-exploit probability distribution group is how to balance them in the optimization process, which is described in the following contents.
\\
\\
\textbf{Dynamic Controller.} 
An online-learnable controller is proposed to determine whether to accept the exploitation distribution or to introduce the stochastic behavior for further optimization. More specifically, a Bernoulli distribution is determined by the information entropy of exploitation distribution to decide the sampling options. 
%
%
\begin{equation}
\label{equ:6}
 H \sim B\left(N, K\right), K = \frac{\rho}{\rho_{max}},
\end{equation}
where $H$ has the possibility of $K$ to be the exploitation distribution, and the possibility of $1-K$ to be the exploration distribution, $N$ represents the number of times to select $H$.
In our algorithm, the Bernoulli distribution $B$ is dynamic, and the controller could adapt to the appropriate state during the super-BERT optimizing process based on the information entropy $\rho$ of the exploitation distribution as shown in Eq.~\ref{equ:6}. On one hand, when the exploitation distribution $P_{\theta}$ tends to convergence, and the corresponding information entropy will decrease. In this case, the Bernoulli distribution $H$ introduces a penalty to the local exploration, i.e., the possibility of sampling from the entire training space will be enhanced. On the other hand, when $P_{\theta}$ has a high information entropy, the distribution $H$ will let the optimization process to exploit in $P_{\theta}$. This simple but effective method could keep a good balance between exploration and exploitation. We demonstrate the effectiveness of our method in Tab.~\ref{tab:4}.
Note that, in the optimization process, the sub-BERTs sampling will only happen on the current training space based on the distributions. 
\subsection{Search For Optimal Architecture}
\label{sec:3.4}
After the optimization process in Sec.~\ref{sec:3.3}, we obtain a probability distribution $\mathcal{P}_{\theta}$ and an optimized super-BERT model. 
There are various works trying to deploy the searched architectures in NAS process to different devices~\cite{cai2018proxylessnas, 49413} .
However, such methods are computationally expensive. Inspired by \cite{tan2019mnasnet}, together with the optimized probability, we design a constraint sensitive reward function Eq.~\ref{equ:7} to further optimize the exploitation distribution $\mathcal{P}_{\theta}$ using stochastic natural gradient optimization in Eq.~\ref{equ:5}, which means $\mathcal{P}_{\theta}$ will concentrate on the sub-BERTs with superior performance under certain constraints. Noted that the optimized weight of super-BERT is fixed in this stage.
\begin{equation}
\label{equ:7}
\begin{aligned}
Reward =
\begin{cases}
\quad \quad \quad ACC\left(m\right), &\mathcal{T}\left(m\right) < {\Omega} \vspace{0.8ex}\\
ACC\left(m\right) \times  \left( \frac{\mathcal{T}\left(m\right)-\Omega}{\mathcal{T}\left({M}\right)-\Omega}  \right) ^{\alpha}, & otherwise
\end{cases}
\end{aligned}
\end{equation}
Given a sub-BERT $m$, $ACC(m)$ denotes its accuracy on the target tasks. The function $\mathcal{T}\left(m\right)$ and $\mathcal{T}\left(M\right)$ measure the related property of sub-BERT $m$ and super-BERT $M$ (\emph{e.g.,} FLOPs, model size). $\Omega$ means the target constraints under certain devices. In our algorithm, we empirically use $\alpha=2$. 
We employ a simple but effective method to search the target sub-BERT. During the searching process, we randomly sample sub-BERTs from the exploitation distribution $\mathcal{P}_{\theta}$, and evaluate using the weight inheriting from the super-BERT. The reward function in Eq.~\ref{equ:6} is used to optimize $\mathcal{P}_{\theta}$. This process will repeat until reaching max searching steps. 
Since we only need to train one super-BERT for various devices, \emph{e.g.,} we could gain the target architecture using our searching algorithm. For $N$ different devices, the training complexity is $\mathcal{O}(1)$.
\begin{algorithm}[!t]
	\caption{Exploit-Explore Stochastic Natural Gradient Optimization}
	\begin{algorithmic}[1]
	\REQUIRE Training Data $D$, Update Interval $I$
	\ENSURE Optimized parameters of $W$ and $\theta$
	\STATE Initialize $W$ and $\theta$; \,  $t \leftarrow 0$;
	\FOR {$t$ = 1,...,$t_{max}$ \text{epoch}}
	    \IF{$t \, \% \, I = 0$ }
	        \STATE Update the controller using Eq.~\ref{equ:6};
	    \ENDIF
	    \STATE Get a batch of samples from $D$;
	    \STATE Decide the distribution by dynamic controller;
	    \STATE Sample $i$ architectures $m_{0},...,m_{i}$ according to the exploitation/exploration distribution;
	    \STATE Compute the loss $\Cost\left(w_{i},m_{i}\right)$ for each $m_{i}$;
	    \STATE Update $\theta$ and $W$ using the gradient in Eq.~\ref{equ:5};
	    \STATE Expand training search space; \, $t \leftarrow t + 1$;
	    \ENDFOR 
	\end{algorithmic}
	\label{alg:1}
\end{algorithm}

\section{Experiments}

In this section, the extensive experiments of YOCO-BERT are conducted against various BERT compression methods. More accurate performance is reported, associating with reduced searching cost. The effectiveness of our method is further demonstrated in ablation analysis.

\textbf{Datasets.} The experiments are conducted on the GLUE benchmark\cite{wang2018glue}, which is widely used to evaluate pre-trained language models. Following BERT~\cite{devlin2019bert}, we do not consider the controversial WNLI dataset.
Regarding SST-2, MNLI, QNLI QQP and RTE, the accuracy is utilized as the metric. While the CoLA is evaluated on matthew's correlation. In terms of MRPC, the F1 and accuracy are evaluated. The average result of MNLI-m and MNLI-mm is reported on MNLI . In particular, the experiments conducted on Fig.~\ref{fig:2} utilize the accuracy as the metric for all datasets. In this paper, if not specifically mentioned, the experimental data is the development set of GLUE by default.

\textbf{Experimental Setup.}
For training the super-BERT, the initial learning rates are $1\times10^{-5}$ for MNLI and QNLI and $2\times10^{-5}$ for the other datasets of GLUE respectively. 
Super-BERT is trained from 1 to 10 epochs according to the dataset size. 
The Adam optimizer is utilized for weight optimization scheduled by a linear annealing. 
Weight is not decay during training. The epsilon for adam is set to be $1\times10^{-8}$. 
Since fine-tuning on BERT easily destroys the pre-trained weight if a high learning rate is set, we perform a complete learning rate decay procedure, in which the learning rate decays to 0. For the most tasks, our total training epoch does not exceed 20 except QNLI and MRPC to avoid over-training. All our experiments are conducted on one Nvidia V100 16GB GPU.

\textbf{Search Space Setups.} To support a scalable super-BERT, our architectures consist of a stack of transformer encoder layers, searching the layer number $d \in \{6,8,10,12\}$. For each layer, we sample from the intermediate size $k \in \{512,768,1024,3072\}$ and head number $h \in \{4,8,12\}$. In this case, we construct a huge search space with $10^{13}$ architectures.

\textbf{Supe-BERT Setups.}
The initial super-BERT weight is directly inherited from the pre-trained $\text{BERT}_{\text{base}}$ model ($d = 12$, $k = 3,072$ and $h = 12$), which composed of 12 identical transformer encoder layers. To speed up the performance evaluation of the compressed model, we directly conducted training process on super-BERT. 

\subsection{Results on GLUE benchmark}
\textbf{Results Under Different Constraints.}
We first compare the performance of the searched sub-BERTs with several constraints against BERT~\cite{devlin2019bert}, where the results are reported in Tab.~\ref{tab:1}. Compared to $\text{BERT}_{\text{base}}$, architectures compressed by YOCO-BERT achieve superior performance with reduced model sizes on most datasets. In particular, $\text{Searched}_{\text{C}}$ achieves $10\%$ accuracy increasing with $64\%$ parameter size compressing on COLA. 
Moreover, most architectures searched by YOCO-BERT outperforms $\text{BERT}_{\text{base}}$ while maintaining a more compact size. 
This further demonstrates the necessity of BERT compressing, as redundant parameters of BERT clearly resulting in significant resource cost~\cite{michel2019sixteen}.

\begin{table}[H]
\vspace{-0.1in}
\caption{Development set results of the GLUE benchmark on the optimal architectures searched by YOCO-BERT in different parameter constraints.}
\label{tab:1}
  \centering
  \renewcommand\arraystretch{1.0}
\begin{tabular}{c|c|ccccccc}
\hline
\textbf{Model Name} & \textbf{Params}    & \textbf{SST-2}     & \textbf{MRPC} & \textbf{CoLA} & \textbf{RTE}  &  \textbf{MNLI} & \textbf{QQP} & \textbf{QNLI}  \\ \hline
$\text{BERT}_{base}$  & 110m  & 92.7          & 89.5           & 54.3        & 71.1        &   83.5    &     89.8      & 91.2   \\ 
\hline
$\text{Searched}_{A}$ & 20m $\sim$ 40m & 84.3 & 81.2 & 15.2 & 65.0 & 71.8 & 88.8 & 69.8 \\
$\text{Searched}_{B}$ & 40m $\sim$ 60m & 92.1 & 88.5  & 55.6 & 69.3 & 81.7 & 89.9 & 85.1   \\
$\text{Searched}_{C}$ & 60m $\sim$ 80m & 92.8 & \textbf{90.3} & 59.8 & 72.9  & 82.6 & \textbf{90.5} & 87.2 \\
$\text{Searched}_{D}$ & 80m $\sim$ 100m & \textbf{93.6} & 89.5 & \textbf{61.3} & \textbf{74.4} & \textbf{83.8} & 90.4  &\textbf{87.5}  \\
\hline
\end{tabular}

\vspace{-0.1in}
\end{table}
\textbf{Comparisons with Previous Compression Work.} We compare the performance and parameters of YOCO-BERT against existing BERT compression methods, including BERT-PKD \cite{sun2019patient}, DistilBERT \cite{sanh2019distilbert}, DynaBERT\cite{hou2020dynabert}, BERT-of-Theseus \cite{xu-etal-2020-bert}, PD-BERT \cite{pd}, MINILM \cite{michel2019sixteen} and LayerDrop \cite{fan2019reducing}. As fine-tuning and data augmentation (DA) are not utilized on YOCO-BERT, NAS-BER\cite{xu2021taskagnostic} is not compared in our experiments. 
For a fair comparison, the model accuracy is compared maintaining a similar FLOPs and parameters. To compare the results of methods that have released the code, we performance the experiments by running the official repository with the same random seed. Otherwise, the results are obtained from the original paper. The corresponding results reported in Tab.~\ref{tab:2}.
In evidence that YOCO-BERT outperforms all previous STOA models compression methods on almost all datasets with a same level constraint (67M). Notably, YOCO-BERT reports a $14.4\%$ higher accuracy on CoLA and a $7.7\%$ improvement on RTE compared with LayerDrop. In terms of  DistilBERT\cite{sanh2019distilbert} which used large external data, the average accuracy of our method outperforms it by $3.7\%$.

\begin{figure*}[t]
\centering
\includegraphics[width=0.9\linewidth]{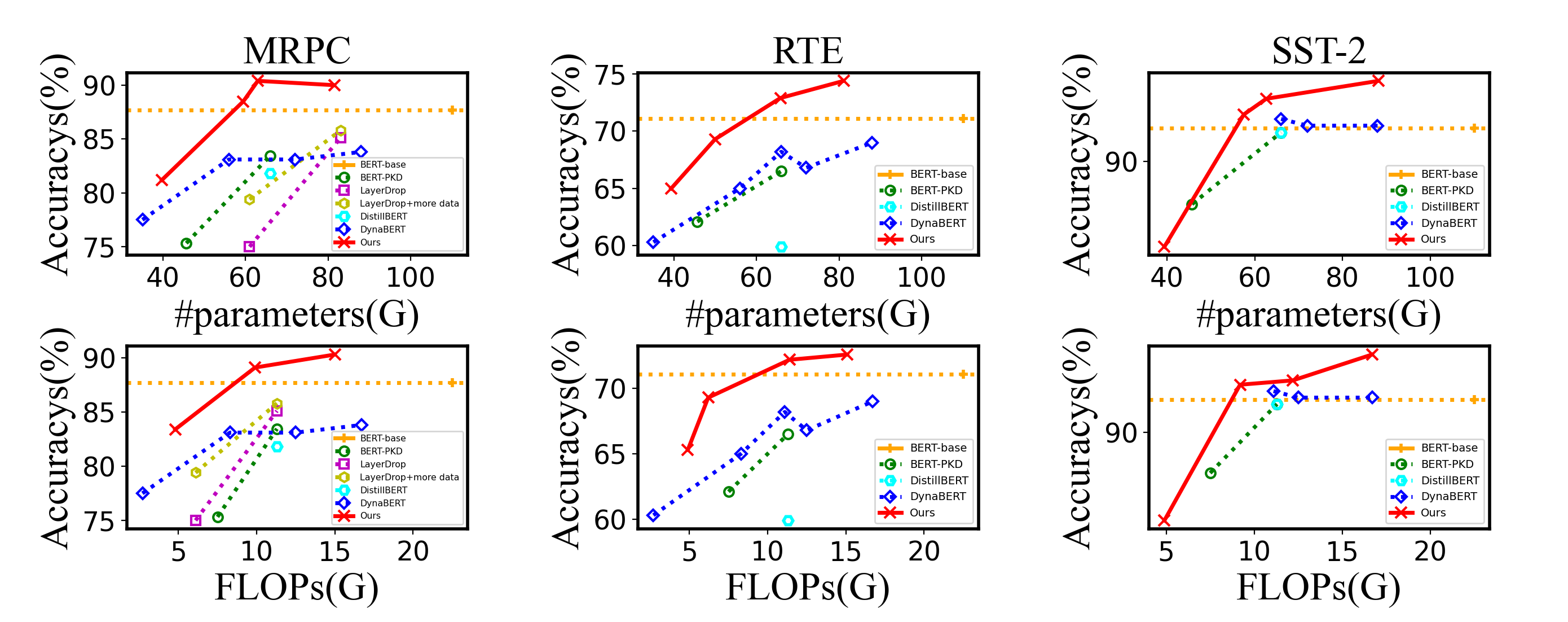}
\caption{Comparison of parameters, FLOPs between YOCO-BERT and previous works under different datasets. We use accuracy as the metric for the three datasets. Since YOCO-BERT does not use external data, Dyanbert’s results are obtained without using data augmentation (DA).}
\label{fig:2}
\end{figure*}

We further demonstrate the effectiveness of YOCO-BERT under different constraints as shown in Fig.~\ref{fig:2}. The comparisons demonstrate that architectures obtained by YOCO-BERT achieve better trade-off between accuracy and model size. Compared with DynaBERT\cite{hou2020dynabert}, YOCO-BERT achieves $6.2\%$, $5.4\%$ accuracy improvement on MRPC and RTE with more compact models respectively. Moreover, YOCO-BERT achieves a $0.5\%$ accuracy improvement with a $53\%$ parameter size reduction on SST-2.
Regarding different datasets, the training time span from 30 minutes (MRPC) to 20 hours (QNLI) depending on the size of the datasets. Compared with DistilBERT\cite{sanh2019distilbert} which costs 720 GPU hours to train, our algorithm is extremely efficient.

\begin{table}[t]
\vspace{-0.1in}
\renewcommand\arraystretch{1.0}
\setlength{\tabcolsep}{0.6 mm}
\caption{Comparision of YOCO-BERT with previous works under the same constraints on the development set of GLUE.}
\label{tab:2}
\centering
\begin{tabular}{c|c|cccccccc}
\hline
\textbf{Method} & \textbf{Params}    & \textbf{SST-2}     & \textbf{MRPC} & \textbf{CoLA}   &  \textbf{RTE} & \textbf{MNLI} & \textbf{QQP} & \textbf{QNLI} & \textbf{Average} \\ \hline
$\text{BERT}_{12}$    & 110m    & 92.7          & 89.5           & 54.3        & 71.1        &   83.5    &    89.8       &  91.2  &  81.7\\
\hline
LayerDrop\cite{fan2019reducing} & 67m & 90.7 & 85.9 & 45.4 & 65.2 & 80.7 & 88.3 & 88.4  & 77.8 \\
DistilBERT\cite{sanh2019distilbert} & 67m & 91.3 & 87.5 & 51.3 & 59.9 & 82.2 & 88.5 & 89.2 & 78.6 \\
BERT-PKD\cite{sun2019patient}  & 67m & 91.3 & 85.7 & 45.5 & 66.5 & 81.3 & 88.4  & 88.4 & 78.2\\
PD-BERT\cite{pd} & 67m & 91.1 & 87.2 & - & 66.7 & 82.5 & 89.1 & 89.0 & - \\
BERT-of-Theseus\cite{xu-etal-2020-bert} & 67m & 91.5 & 89.0 & 51.1 & 68.2 & 82.3 & 89.6 & \textbf{89.5} & 80.2 \\
MINILM\cite{michel2019sixteen} & 67m & 92.0 & 88.4 & 49.2 & 71.5 & - & \textbf{91.0} & 91.0 & -  \\
\hline
\textbf{YOCO-BERT} & \textbf{59m$\sim$67m} & \textbf{92.8} & \textbf{90.3} & \textbf{59.8} & \textbf{72.9} & \textbf{82.6} & 90.5  & 87.2  & \textbf{82.3} \\
\hline
\end{tabular}
\end{table}

\textbf{Comparing with HAT\cite{hanruiwang2020hat} and OFA\cite{Cai2020Once-for-All:}.} 
HAT\cite{hanruiwang2020hat} demonstrate a transformer-based compression on machine translation tasks. OFA\cite{Cai2020Once-for-All:} compresses CNN-based model elastically on a wide range of devices. For a fair comparison, we re-implement these two methods and search optimal architectures in our search space. The results are reported in Tab.~\ref{tab:3}. As we can see, YOCO-BERT outperforms HAT\cite{hanruiwang2020hat} and OFA\cite{Cai2020Once-for-All:} on all datasets in achieving higher accuracy on both compression ratios. Regarding $0.75$ compression ratio on RTE dataset, YOCO-BERT reports 10.1$\%$ and 12.3$\%$ improvements for HAT and OFA respectively. Since HAT and OFA require additional fine-tuning process, there is a linear growing of the total costs concerning these methods, preventing the directly migration of these methods. In contrast to these methods, the weight is directly inherited from super-BERT in our method, reducing the cost of specialized BERT compression from $\mathcal{O}(N)$ to $\mathcal{O}(1)$.


\begin{table}[H]
\vspace{-0.1in}
\renewcommand\arraystretch{1.0}
\caption{Comparison with HAT\cite{hanruiwang2020hat} and OFA\cite{Cai2020Once-for-All:} }
\label{tab:3}
\centering
\begin{tabular}{l|cc|cc|cc}
\hline
& \multicolumn{2}{c|}{\textbf{MRPC}} & \multicolumn{2}{c|}{\textbf{SST-2}} & \multicolumn{2}{c}{\textbf{RTE}}  \\ \hline
\textbf{Compression Ratio}             & 0.75x &        0.5x         & 0.75x &         0.5x         & 0.75x &        0.5x           \\ \hline
$\text{HAT-BERT}$  &  82.2 &82.6               & 88.6  &    88.6               & 65.0 & 64.6        \\ \hline
$\text{OFA-BERT}$   &  87.6 & 85.2                  & 89.3  & 89.8                  & 62.8 & 65.3    \\ \hline
$\textbf{YOCO-BERT}$ & \textbf{90.4} &   \textbf{87.6}                &  \textbf{92.9} & \textbf{91.9}                  & \textbf{75.1} & \textbf{69.3}     \\ 
\hline
\end{tabular}
\vspace{-0.1in}
\end{table}

\subsection{Ablation Study}
\textbf{Impact of Exploit-Explore Stochastic Natural Gradient Optimization.}
To analyse the effectiveness of EE-SNG optimization algorithm, three experiments are conducted under different constraints: (a) Exploit-Only BERT compression; (b) Explore-Only BERT compression; (c) YOCO-BERT. The results are reported in Tab.~\ref{tab:4}, compared to (a) and (b), YOCO-BERT provides better performance. When compressing BERT with $0.75x$ compression ratio, YOCO-BERT outperforms Explore-Only 9.8$\%$ accuracy on RTE, while 5.3$\%$ accuracy increasing is achieved on MRPC. This proves the importance of balancing exploration and exploitation.
\begin{table}[H]
\vspace{-0.1in}
\caption{Ablation study of the Exploit-Explore Stochastic Natural Gradient Optimization} 
\renewcommand\arraystretch{1.0}
\label{tab:4}
  \centering
\begin{tabular}{l|cc|cc|cc}
\hline
& \multicolumn{2}{c|}{Exploit-Only} & \multicolumn{2}{c|}{Explore-Only} & \multicolumn{2}{c}{\textbf{YOCO-BERT}}  \\ \hline
\textbf{Compression Ratio}             & 0.75x &        0.5x         & 0.75x &         0.5x         & 0.75x &        0.5x           \\ \hline
$\text{MRPC}$  &  $88.7_{\textcolor{red}{-1.7}}$ & $84.1_{\textcolor{red}{-3.5}}$               &  $85.1_{\textcolor{red}{-5.3}}$ &    $85.5_{\textcolor{red}{-2.1}}$               & \textbf{90.4} & \textbf{87.6}        \\ 
$\text{RTE}$   & $72.5_{\textcolor{red}{-2.6}}$  & $66.4_{\textcolor{red}{-2.9}}$               &  $65.3_{\textcolor{red}{-9.8}}$ & $67.1_{\textcolor{red}{-2.2}}$               & \textbf{75.1} & \textbf{69.3}    \\ \hline
 
\end{tabular}
\vspace{-0.1in}
\end{table}
\begin{wrapfigure}{l}{0pt}
\centering
  \renewcommand\arraystretch{1.0}
\includegraphics[width=.5\textwidth]{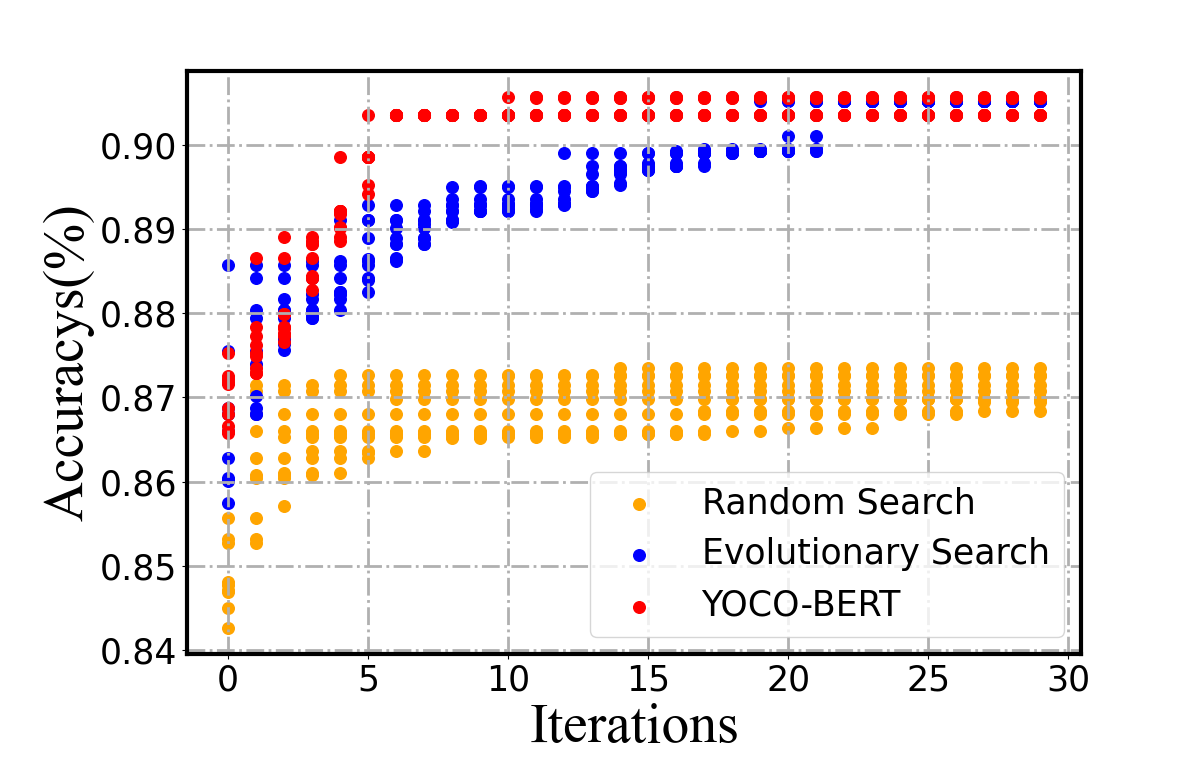}
\caption{Comparison of evolutionary search, random search and YOCO-BERT in the search stage.\label{fig:3}}
\vspace{-0.20in}
\end{wrapfigure}
\textbf{Impact of the Searching Algorithm.}
To analyse the impact of the probability distribution-based search algorithm, we compared our algorithm against random search and evolutionary search on the MRPC dataset, while other experimental conditions are fixed. Results are shown in Fig.~\ref{fig:3}, ten optimal sub-architectures are recorded every steps in each algorithms. Our algorithm outperforms all other methods in achieving both highest accuracy and convergent rate. In fact, the probability distribution has been well optimized during the super-BERT training process. Therefore, the optimal architecture could be identified more efficiently from optimized distribution during the searching process.

\textbf{Fine-tuning Searched Architectures.}
In the previous experiments, we reported the results without fine-tuning stage, \emph{i.e.} directly inheriting the weights from super-BERT and evaluating. In this section, we explore the impact of the fine-tuning to the searched architectures.  
Fine-tuning is performed regarding different learning rates for additional ten epochs on the three datasets (MRPC, SST-2, RTE). As shown in Tab.~\ref{tab:5}, the fine-tuning stage obviously is not able to capture a significant improving and even acts a negative impact on the sampled sub-BERTs. In NAS, this phenomenon has been mentioned in BigNAS\cite{49413} as well.

\begin{table}[H]
\vspace{-0.1in}
\caption{Ablation study of the fine-tuning process for sampled architecture from super-BERT model. We compare the performance of sub-architectures fine-tuned with various learning rates (FT at different LR) and directly inheriting the weight from super-BERT (w/o Fine-tuning).}
\label{tab:5}
 \centering
 \renewcommand\arraystretch{1.0}
\begin{tabular}{c|c|c|c|c}
\hline
\multirow{2}{*}{Dataset} & \textbf{FT} & \textbf{FT} & \textbf{FT} & \multirow{1}{*}{\textbf{w/o}} \\
 & lr = $2*e^{-5}$ & lr = $2*e^{-6}$ & lr = $2*e^{-7}$ & Finetuning \\
\hline
MRPC & $85.7_{\textcolor{red}{-4.6}}$ & $88.3_{\textcolor{red}{-2.0}}$ & $89.4_{\textcolor{red}{-0.9}}$ & \textbf{90.3}    \\
SST-2 & $89.2_{\textcolor{red}{-3.6}}$  &  $91.3_{\textcolor{red}{-1.5}}$ & $92.7_{\textcolor{red}{-0.1}}$ & \textbf{92.8}  \\
RTE & $67.9_{\textcolor{red}{-5.0}}$ & $67.5_{\textcolor{red}{-5.4}}$ & $72.2_{\textcolor{red}{-0.7}}$ & \textbf{72.9}\\
\hline
\end{tabular}
\vspace{-0.1in}
\end{table}


\section{Conclusion}
In this paper, we propose a BERT compression algorithm, YOCO-BERT, that automatically compresses transformer-based models to satisfy resource constraints in an elastic and effective way.
We first perform NAS in a huge search space with $10^{13}$ architectures within the BERT family. Under the guidance of the probability distribution, the training is conducted in a progressive way such that the sub-BERTs sampled from the final super-BERT model are able to provide good performance.
Then we use a lightweight search based on a probability distribution to obtain the optimal model.
In this way, \textbf{we only need to compress once for different hardware constraints}.
Extensive experiments on GLUE demonstrate that YOCO-BERT shows better accuracy and compression ratio compared with previous compression methods.
Future works may include extending the approach for more tasks and broader models.


\renewcommand\refname{References}
\bibliographystyle{plain}
\bibliography{paper}

\end{document}